\documentclass[11pt]{article}
\usepackage{coling2016}
\usepackage{times}
\usepackage{url}
\usepackage{latexsym}

\usepackage{float}
\usepackage{multirow}
\usepackage{multicol}
\usepackage{wrapfig}
\usepackage{subfig}
\usepackage{booktabs}

\usepackage{graphicx}
\usepackage{stfloats}
\usepackage{caption}
\usepackage{amsmath}
\usepackage{lipsum}

\usepackage{devanagari}
\usepackage{pifont}
\usepackage{bbding}

\newcommand*{\affaddr}[1]{#1} 
\newcommand*{\affmark}[1][*]{\textsuperscript{#1}}
\newcommand*{\email}[1]{\texttt{#1}}

\newcommand*\OK{\ding{51}}

\title{Towards Sub-Word Level Compositions for\\ Sentiment Analysis of Hindi-English Code Mixed Text}

\begin{document}

\author{%
 Ameya Prabhu\footnotemark[1] \affmark[1] ,Aditya Joshi  \thanks{* indicates these authors contributed equally to this work.} \affmark[2], Manish Shrivastava\affmark[3] and Vasudeva Varma\affmark[2]
\\ 
\affaddr{\affmark[1]Centre for Visual Information Technology}\\
\affaddr{\affmark[2]Search and Information Extraction Lab}\\
\affaddr{\affmark[3]Language Technologies Research Center}\\
\affaddr{International Institute of Information Technology, Hyderabad (India)}\\
\email{
\{ameya.prabhu,aditya.joshi\} @research.iiit.ac.in}\\
\email{
\{m.shrivastava, vv\} @.iiit.ac.in}%
}
\
\maketitle
\begin{abstract}
Sentiment analysis (SA) using code-mixed data from social media has several applications in opinion mining ranging from customer satisfaction to social campaign analysis in multilingual societies. Advances in this area are impeded by the lack of a suitable annotated dataset. We introduce a Hindi-English (Hi-En) code-mixed dataset for sentiment analysis and perform empirical analysis comparing the suitability and performance of various state-of-the-art SA methods in social media.

In this paper, we introduce learning sub-word level representations in LSTM (Subword-LSTM) architecture instead of character-level or word-level representations. This linguistic prior in our architecture enables us to learn the information about sentiment value of important morphemes. This also seems to work well in highly noisy text containing misspellings as shown in our experiments which is demonstrated in morpheme-level feature maps learned by our model. Also, we hypothesize that encoding this linguistic prior in the Subword-LSTM architecture leads to the superior performance. Our system attains accuracy 4-5\% greater than traditional approaches on our dataset, and also outperforms the available system for sentiment analysis in Hi-En code-mixed text by 18\%.
\end{abstract}

\section{Introduction}
\label{sec:intro}

\blfootnote{
    \hspace{-0.65cm}  
    This work is licensed under a Creative Commons Attribution 4.0 International Licence.\\ Licence details: http://creativecommons.org/licenses/by/4.0/
}

Code Mixing is a natural phenomenon of embedding linguistic units such as phrases, words or morphemes of one language into an utterance of another \cite{Muysken:2000,Duran:1994,Gysels:1994}. Code-mixing is widely observed in multilingual societies like India, which has 22 official languages most popular of which are Hindi and English. With over 375 million Indian population online, usage of Hindi has been steadily increasing on the internet.

This opens up tremendous potential for research in sentiment and opinion analysis community for studying trends, reviews, events, human behaviour as well as linguistic analysis. Most of the current research works have involved sentiment polarity detection \cite{Feldman:2013,Liu:2012,Pang:2008} where the aim is to identify whether a given sentence or document is (usually) positive, negative or neutral. Due to availability of large-scale monolingual corpora, resources and widespread use of the language, English has attracted the most attention. 

Seminal work in sentiment analysis of Hindi text was done by \newcite{Joshi:2010} in which the authors built three step fallback model based on classification, machine translation and sentiment lexicons. They also observed that their system performed best with unigram features without stemming.  \newcite{Bakliwal:2012} generated a sentiment lexicon for Hindi and validated the results on translated form of Amazon Product Dataset \newcite{Blitzer:07}. \newcite{Das:10} created Hindi SentiWordNet, a sentiment lexicon for Hindi.

Sentiment Analysis in Code-mixed languages has recently started gaining interest owing to the rising amount of non-English speaking users. \newcite{Sharma:2015} segregated Hindi and English words and calculated final sentiment score by lexicon lookup in respective sentient dictionaries.

Hindi-English (Hi-En) code mixing allows ease-of-communication among speakers by providing a much wider variety of phrases and expressions. A common form of code mixing is called as \textit{romanization} \footnote{https://en.wikipedia.org/wiki/Romanization}, which refers to the conversion of writing from a different writing system to the Roman script. But this freedom makes the task for developing NLP tools more difficult, highlighted by \cite{Chittaranjan:2014,Vyas:2014,Barman:2014}. Initiatives have been taken by shared tasks \cite{sequiera:15,solorio:14}, however they do not cover the requirements for a sentiment analysis system.

Deep learning based approaches \cite{Zhang:15,Socher:2013} have been demonstrated to solve various NLP tasks. We believe these can provide solution to code-mixed and romanized text from various demographics in India, as similar trends are followed in many other Indian languages too. \newcite{Santos:14} demonstrated applicability of character models for NLP tasks like POS tagging and Named Entity Recognition \cite{Santos:15}. LSTMs have been observed to outperform baselines for language modelling \cite{Kim:2015} and classification \cite{zhou:15}. In a recent work, \cite{bojan16} proposed a skip-gram based model in which each word is represented as a bag of character n-grams. The method produced improved results for languages with large vocabularies and rare words.


The \textit{romanized} code mixed data on social media presents additional inherent challenges such as contractions like "between" $\rightarrow$ "btwn", non-standard spellings such as "cooolll" or "bhut bdiya" and non-grammatical constructions like "sir hlp plzz naa". Hindi is phonetically typed while English (Roman script) doesn't preserve phonetics in text. Thus, along with diverse sentence construction, words in Hindi can have diverse variations when written online, which leads to large amount of tokens, as illustrated in Table \ref{tab:spellvariations}. Meanwhile there is a lack of a suitable dataset. 

\begin{table}[t]
\centering
\makebox[0pt][c]{\parbox{\textwidth}{%
    \begin{minipage}[t]{0.4\hsize}\centering
      \begin{tabular}{l}
      \hline
      {\bf Sentence variations}\\
      \hline
          Trailer dhannnsu hai bhai\\
          Dhannnsu trailer hai bhai\\
          Bhai trailer dhannnsu hai\\
          Bhai dhannnsu trailer hai\\
      \hline
      \end{tabular}
        \caption{Illustration of free structure present in code mixed text. All sentences convey the same meaning.}
        \label{tab:sentencelevelvariations}
    \end{minipage}
    \hfill
    \begin{minipage}[t]{0.55\hsize}\centering
      \begin{tabular}{ p{1.8cm} | c | p{4cm} }
      \hline
      {\bf Word} & {\bf Meaning} & {\bf Appearing Variations}\\ 
      \hline
          {\dn b\7ht} (bahut)  & very & bahout bohut bhout bauhat bohot bahut bhaut bahot bhot  \\
    {\dn \7mbArk} (mubaarak) & wishes & mobarak mubarak mubark \\
    {\dn \6pyAr} (pyaar) & love & pyaar peyar pyara piyar pyr piyaar pyar \\
      \hline
      \end{tabular}
        \caption{Spelling variations of \textit{romanized} words in our Hi-En code-mix dataset.}
        \label{tab:spellvariations}
    \end{minipage}
    \hfill
}}
\end{table}

Our contributions in this paper are (i) Creation, annotation and analysis of a Hi-En code-mixed dataset for the sentiment analysis, (ii) Sub-word level representations that lead to better performance of LSTM networks compared to Character level LSTMs (iii) Experimental evaluation for suitability and evaluation of performance of various state-of-the-art techniques for the SA task, (iv) A preliminary investigation of embedding linguistic priors might be encoded for SA task by char-RNN architecture and the relation of architecture with linguistic priors, leading to the superior performance on this task.\\
Our paper is divided into the following sections: \\
We begin with an introduction to Code Mixing and romanization in Section \ref{sec:intro}. We mention the issues with code-mixed data in context of Sentiment Analysis and provides an overview of existing solutions. We then discusses the process of creation of the dataset and its features in Section \ref{sec:dataset}. In Section \ref{sec:theory}, we introduce Sub-word level representation and explains how they are able to model morphemes along with propagating meaningful information, thus capturing sentiment in a sentence. Then in Section \ref{sec:experiments}, we explain our experimental setup, describe the performance of proposed system and compare it with baselines and other methods, proceeded by a discussion on our results. 
\section{Dataset} \label{sec:dataset}
We collected user comments from public Facebook pages popular in India. We chose pages of Salman Khan, a popular Indian actor with massive fan following, and Narendra Modi, the current Prime Minister of India. The pages have 31 million and 34 million facebook user likes respectively. These pages attract large variety of users from all across India and contain lot of comments to the original posts in code-mixed representations in varied sentiment polarities. We manually pre-processed the collected data to remove the comments that were not written in roman script, were longer than 50 words, or were complete English sentences. We also removed the comments that contained more than one sentence, as each sentence might have different sentiment polarity.
Then, we proceeded to manual annotation of our dataset. The comments were annotated by two annotators in a 3-level polarity scale - positive, negative or neutral. Only the comments with same polarity marked by both the annotators are considered for the experiments. They agreed on the polarity of 3879 of 4981 (77\%) sentences. The Cohen's Kappa coefficient \cite{Cohen:60} was found to be 0.64. We studied the reasons for misalignment and found that causes typically were due to difference in perception of sentiments by individuals, different interpretations by them and sarcastic nature of some comments which is common in social media data. The dataset contains 15\% negative, 50\% neutral and 35\% positive comments owing to the nature of conversations in the selected pages.

The dataset exhibits some of the major issues while dealing with code-mixed data like short sentences with unclear grammatical structure. Further, \textit{romanization} of Hindi presents an additional set of complexities due to loss of phonetics and free ordering in sentence constructions as shown in Table \ref{tab:sentencelevelvariations}. This leads to a number of variations of how words can be written.  Table \ref{tab:spellvariations} contains some of the words with multiple spelling variations in our dataset, which is one of the major challenges to tackle in Hi-En code-mixed data.

\begin{table}[h]
\centering
\begin{tabular}{l|c|c|c|c|c}
\hline
  {\bf Dataset} & {\bf Size} &{\bf \# Vocab} &{\bf Social} &{\bf CM} &{\bf Sentiment} \\ 
  \hline
  STS-Test & 498 & 2375 & \OK &   & \OK  \\
  OMD  & 3238  & 6211 & \OK &   & \OK\\
  SemEval'13  & 13975  & 35709  & \OK &  & \OK\\
  IMDB & 50000 & 5000 &  &  & \OK\\
  \cite{Vyas:2014} & 381 & - & \OK & \OK &  \\
  \hline
  Ours & 3879 & 7549 & \OK & \OK & \OK \\
  \hline
\end{tabular}
\caption{Comparison with other datasets.}
\label{tab:datasetcomparison}
\end{table} 

Popular related datasets are listed in Table \ref{tab:datasetcomparison}. STS, SemEval, IMDB etc. have been explored for SA tasks but they contain text in English. The dataset used by \newcite{Vyas:2014} contains Hi-En Code Mixed text but doesn't contain sentiment polarity. We constructed a code mixed dataset with sentiment polarity annotations, and the size is comparable with several datasets. Table \ref{tab:examples} shows some examples of sentences from our dataset. Here, we have phrases in Hindi (source language) written in English (target) language. 

\begin{table}[h]
\resizebox{\textwidth}{!}{%
    \begin{tabular}{ |l|l|c| }
    \hline
{\bf Example} & {\bf Approximate Meaning} & {\bf Sentiment Polarity}\\
    \hline
Aisa PM naa hua hai aur naa hee hoga & Neither there has been a PM like him, nor there will be & Positive\\
abe kutte tere se kon baat karega & Who would talk to you, dog? & Negative\\
Trailer dhannnsu hai bhai & Trailer is awesome, brother. & Positive\\
    \hline
    \end{tabular}}
    \caption{Examples of Hi-En Code Mixed Comments from the dataset.}
    \label{tab:examples}
\end{table} 
Our dataset and code is freely available for download \footnote{https://github.com/DrImpossible/Sub-word-LSTM} to encourage further exploration in this domain.
\clearpage
\section{Learning Compositionality}\label{sec:theory}
Our target is to perform sentiment analysis on the above presented dataset. Most commonly used statistical approaches learn word-level feature representations. We start our exploration for suitable algorithms from models having word-based representations. 
\subsection{Word-level models}

Word2Vec\cite{mikolov:13} and Word-level RNNs (Word-RNNs) \cite{Luong:13} have substantially contributed to development of new representations and their applications in NLP such as in Summarization \cite{cao:2015} and Machine Translation \cite{cho:14}. They are theoretically sound since language consists of inherently arbitrary mappings between ideas and words. Eg: The words person(English) and insaan(Hindi) do not share any priors in their construction and neither do their constructions have any relationship with the semantic concept of a person. Hence, popular approaches consider lexical units to be independent entities. However, operating on the lexical domain draws criticism since the finite vocabulary assumption; which states that models assume language has finite vocabulary but in contrast, people actively learn \& understand new words all the time.

Excitingly, our dataset seems suited to validate some of these assumptions. In our dataset, vocabulary sizes are greater than the size of the dataset as shown in Table  \ref{tab:datasetcomparison}. Studies on similar datasets have shown strong correlation between number of comments and size of vocabulary \cite{saif:13}. This rules out methods like Word2Vec, N-grams or Word-RNNs which inherently assume a small vocabulary in comparison to the data size. The finite vocabulary generally used to be a good approximation for English, but is no longer valid in our scenario. Due to the high sparsity of words themselves, it is not possible to learn useful word representations. This opens avenues to learn non-lexical representations, the most widely studied being character-level representations, which is discussed in the next section.
\subsection{Character-level models}
Character-level RNNs (Char-RNNs) have recently become popular, contributing to various tasks like \cite{Kim:2015}. They do not have the limitation of vocabulary, hence can freely learn to generate new words. This freedom, in fact, is an issue: Language is composed of lexical units made by combining letters in some specific combinations, i.e. most of the combinations of letters do not make sense. The complexity arises because the mappings between meaning and its construction from characters is arbitrary.  Character models may be apriori inappropriate models of language as characters individually do not usually provide semantic information. For example, while ``
$King-Man + Women = Queen$'' is semantically interpretable by a human, 
 ``$Cat - C+B = Bat$'' lacks any linguistic basis.
 
But, groups of characters may serve semantic functions. This is illustrated by $Un + Holy = Unholy$ or $Cat + s = Cats$ which is semantically interpretable by a human.
Since sub-word level representations can generate meaningful lexical representations and individually carry semantic weight, we believe that sub-word level representations consisting composition of characters might allow generation of new lexical structures and serve as better linguistic units than characters.
\subsection{Sub-word level representations}
\begin{figure*}[t]
\centering
\captionsetup{justification=centering}
    \includegraphics[width=11cm,height=8cm]{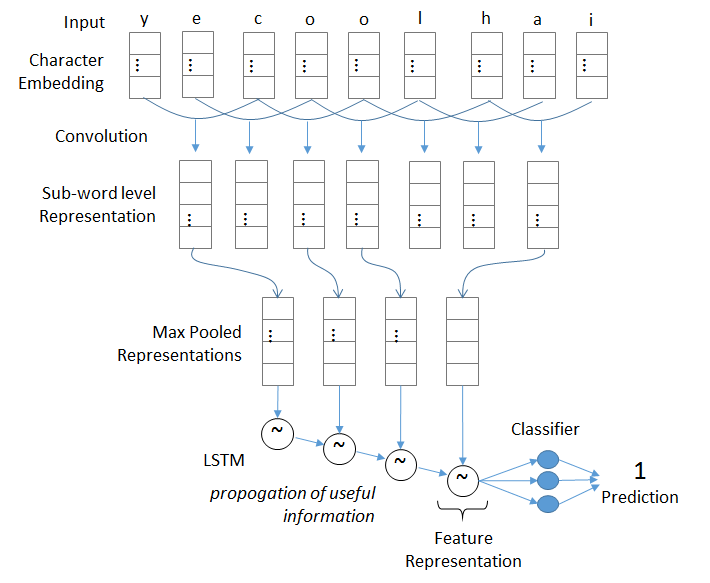}
    \caption{Illustration of the proposed methodology}
    \label{fig:methodology}
\end{figure*}

Lexicon based approaches for the SA task \cite{Taboada:2011,Sharma:2015} perform a dictionary look up to obtain an individual score for words in a given sentence and combine these scores to get the sentiment polarity of a sentence. We however want to use intermediate sub-word feature representations learned by the filters during convolution operation. Unlike traditional approaches that add sentiment scores of individual words, we propagate relevant information with LSTM and compute final sentiment of the sentence as illustrated in Figure \ref{fig:methodology}.\\
\textbf{Hypothesis:} We propose that incorporating sub-word level representations into the design of our models should result in better performance. This would also serve as a test scenario for the broader hypothesis proposed by Dyer et. al. in his impressive ICLR keynote \footnote{Available at: http://videolectures.net/iclr2016\_dyer\_model\_architecture/ } - Incorporating linguistic priors in network architectures lead to better performance of models.\\
\textbf{Methodology:} We propose a method of generating sub-word level representations through 1-D convolutions on character inputs for a given sentence. Formally, let $C$ be the set of characters and $T$ be an set of input sentences. The sentence $s \in T$  is made up of a sequence of characters $[c_1,....,c_l]$ where $l$ is length of the input.

Hence, the representation of the input $s$ is given by the matrix $Q \in R^{d\times l}$ where $d$ is the dimensionality of character embedding that corresponding to $[c_1,....,c_l]$. We perform convolution of $Q$ with a filter $H \in R^{d\times m}$ of length $m$ after which we add a bias and apply a non-linearity to obtain a feature map $f \in R^{l-m+1}$. Thus we can get  sub-word level (morpheme-like) feature map. Specifically, the $i^{th}$ element of $f$ is given by:
\begin{align}
  f[i] &= g( ( Q [:, i:i+m-1] \ast H ) +b   ) 
\end{align}
where $Q[:, i:i+m-1]$ is the matrix of $(i)^{th}$ to $(i+m-1)^{th}$ character embedding and $g$ corresponds to ReLU non-linearity. \\
Finally, we pool the maximal responses from $p$ feature representations corresponding to selecting sub-word representations as: 
\begin{align}
  y_{i} &= max (f[p*(i:i+p-1)])
\end{align}
Next, we need to model the relationships between these features $y^i[:]$ in order to find the overall sentiment of the sentence. This is achieved by LSTM\cite{Graves:13lstm} which is suited to learning to propagate and 'remember' useful information, finally arriving at a sentiment vector representation from the inputs. 
We provide $f_t$ as an input to the memory cell at time $t$. We then compute values of   $I_t$ - the input gate, $\tilde{C}_t$ - the candidate value for the state of the memory cell at time $t$ and $f_t$ - the activation of the forget gate, which can be used to compute the information stored in memory cell at time $t$. 
With the new state of memory cell $C_t$, we can compute the output feature representation by:
\begin{align}
O_t &= \sigma (W y_t + U h_(t-1) + V (C_t +b)  \\
h_t &= O_t tanh(C_t)
\end{align}
where $W$,$U$ and $V$ are weight matrices and $b_i$ are biases. After l steps, $h_l$ represents the relevant information retained from the history. That is then passed to a fully connected layer which calculates the final sentiment polarity as illustrated in the Figure \ref{fig:methodology}.

Figure \ref{fig:architecture} gives schematic overview of the architecture. We perform extensive experiments to qualitatively and quantitatively validate the above claims as explained in the next section. \\

\section{Experiments}
\label{sec:experiments}
We perform extensive evaluation of various approaches, starting with a suitability study for the nature of approaches that would be able to generalize to this data. We compare our approaches with the state-of-the-art methods which are feasible to generalize on code-mixed data and \cite{Sharma:2015}, the current state-of-the-art in Hi-En code-mixed SA task.

\subsection{Method Suitability}
Following approaches have been used for performing SA tasks in English but do not suit mix code setting:
\begin{itemize}
    \setlength{\itemsep}{3pt}
    \setlength{\parskip}{0pt}
    \setlength{\parsep}{0pt} 
  \item Approaches involving NLP tools: RNTN \cite{Socher:2013} etc which involve generation of parse trees which are not available for code mixed text;
  \item Word Embedding Based Approaches: Word2Vec, Word-RNN may not provide reliable embedding in situations with small amount of highly sparse dataset.
  \item Surface Feature engineering based approaches: Hashtags, User Mentions, Emoticons etc. may not exist in the data.
\end{itemize}

\begin{figure}
\centering
  \begin{minipage}{.48\textwidth}
  \centering
      \captionsetup{justification=centering}
      \includegraphics[width=5cm]{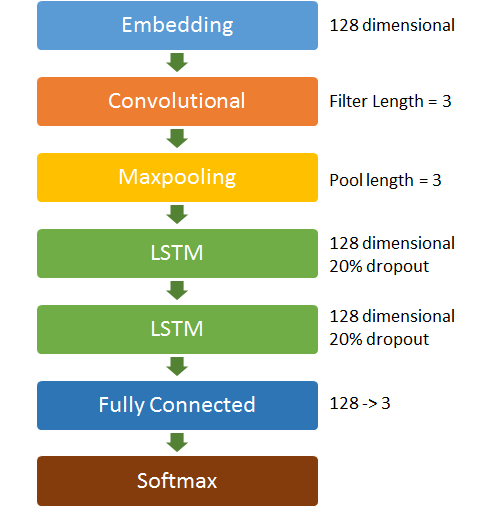}
      \caption{Schematic overview of the architecture.}
      \label{fig:architecture}
    \end{minipage}
    \begin{minipage}{.48\textwidth}
    \centering
      \captionsetup{justification=centering}
      \includegraphics[width=5cm]{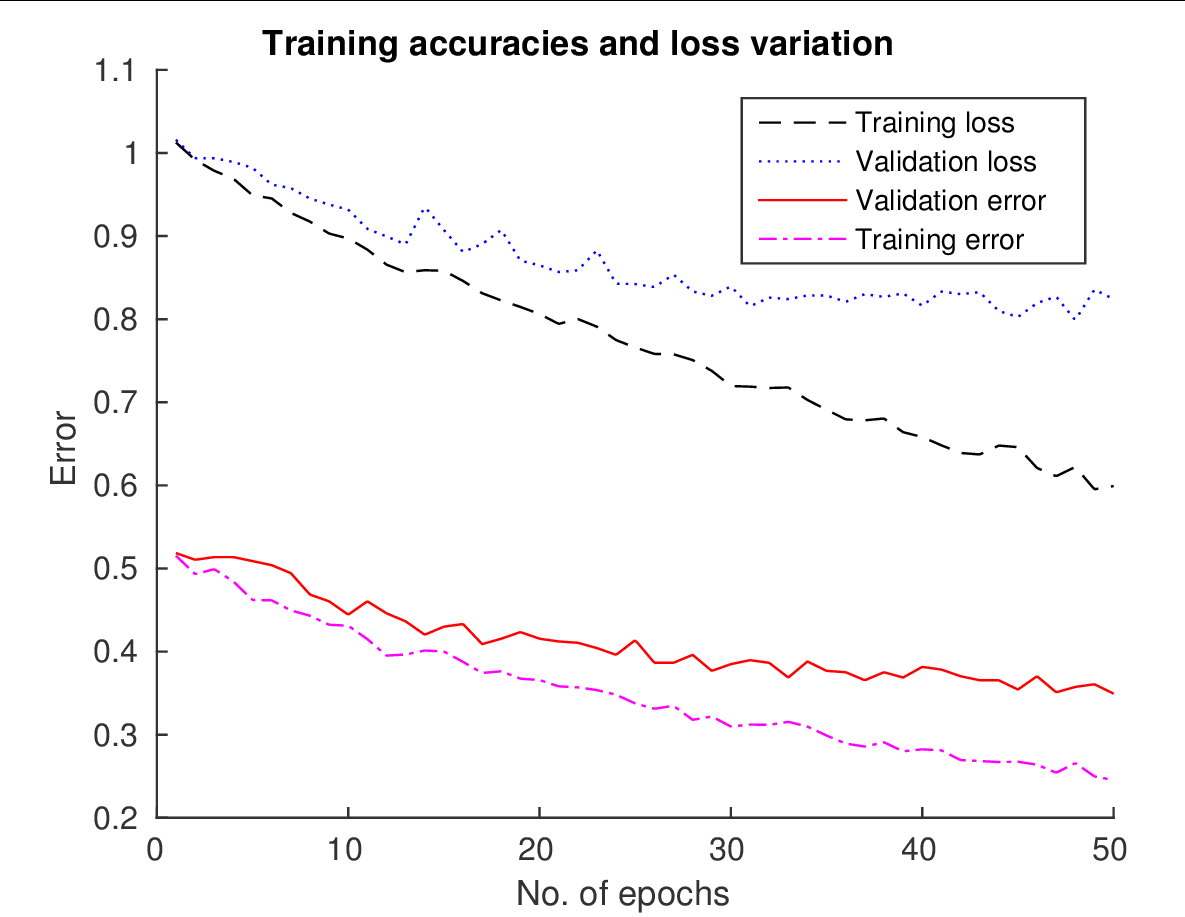}
      \caption{Training accuracy and loss variation.}
      \label{fig:loss}
    \end{minipage}
\end{figure}

\subsection{Experimental Setup}
Our dataset is divided into 3 splits- Training, validation and testing. We first divide the data into randomized 80-20 train test split, then further randomly divide the training data into 80-20 split to get the final training, validation and testing data.

As the problem is relatively new, we compare state of the art sentiment analysis techniques \cite{Wang:2012,Pang:2008} which are generalizable to our dataset. We also compare the results with system proposed by \newcite{Sharma:2015} on our dataset. As their system is not available publicly, we implemented it using language identification and transliteration using the tools provided by \newcite{Bhat:14} for Hi-En Code Mixed data. The polarity of thus obtained tokens is computed from SentiWordNet \cite{Esuli:06} and Hindi SentiWordNet \cite{Das:10} to obtain the polarity of words, which are then voted to get final polarity of the sentence. 

The architecture of the proposed system (Subword-LSTM) is described in Figure \ref{fig:architecture}. We compare it with a character-level LSTM (Char-LSTM) following the same architecture without the convolutional and maxpooling layers. We use Adamax \cite{Kingma:2014} (a variant of Adam based on infinity norm) optimizer to train this setup in an end-to-end fashion using batch size of 128. We use very simplistic architectures because of the constraint on the size of the dataset.  As the datasets in this domain expand, we would like to scale up our approach to bigger architectures. The stability of training using this architecture is illustrated in Figure \ref{fig:loss}.

\begin{table*}[t]
\resizebox{\textwidth}{!}{%
\begin{tabular}{|c|l|c|c|c|c|}
\hline
{\bf Method} & {\bf Reported In} & \multicolumn{2}{|c|}{Our dataset}  &  \multicolumn{2}{|c|}{SemEval' 13} \\
\hline
&&{\bf Accuracy} & {\bf F1-Score} &{\bf Accuracy} & {\bf F1-Score}\\
\hline
NBSVM (Unigram)    & \cite{Wang:2012} & 59.15\% & 0.5335 & 57.89\% & 0.5369 \\
NBSVM (Uni+Bigram) & \cite{Wang:2012} & 62.5\%  & 0.5375 & 51.33\% & 0.5566 \\
MNB (Unigram)      & \cite{Wang:2012} & 66.75\% & 0.6143 & 58.41\% & 0.4689 \\
MNB (Uni+Bigram)   & \cite{Wang:2012} & 66.36\% & 0.6046 & 58.4\%  & 0.469  \\
MNB (Tf-Idf)       & \cite{Wang:2012} & 63.53\% & 0.4783 & 57.82\% & 0.4196 \\
SVM (Unigram)      & \cite{Pang:2008} & 57.6\%  & 0.5232 & 57.6\%  & 0.5232 \\
SVM (Uni+Bigram)   & \cite{Pang:2008} & 52.96\% & 0.3773 & 52.9\%  & 0.3773 \\
\hline
Lexicon Lookup & \cite{Sharma:2015} & 51.15\% & 0.252 & N/A & N/A \\
\hline
Char-LSTM & Proposed & 59.8\% & 0.511 & 46.6\% & 0.332\\
Subword-LSTM & Proposed & \textbf{69.7}\% & \textbf{0.658} & \textbf{60.57\%} & 0.537\\
\hline

\end{tabular}}
\caption{Classification results show that the proposed system provides significant improvement over traditional and state of art method for Sentiment Analysis in Code Mixed Text }
\label{tab:results_codemix}
\end{table*}

\begin{table*}
\centering
\captionsetup{justification=centering}
    \includegraphics[width=\textwidth]{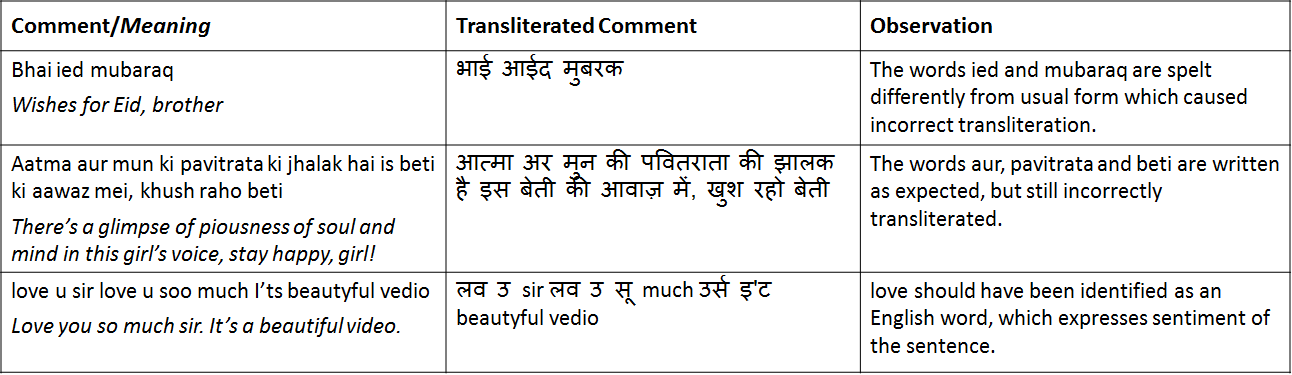}
    \caption{Output produced a by Hi-En Transliteration Tool}
    \label{fig:transliteration}
\end{table*}
\subsection{Observations}
In the comparative study performed on our dataset, we observe that Multinomial Naive Bayes performs better than SVM\cite{Pang:2008} for snippets providing additional validation to this hypothesis given by \newcite{Wang:2012}.

We also observe that unigrams perform better than bigrams and Bag of words performs better than tf-idf in contrast to trends in English, as the approaches inducing more sparsity would yield to poorer results because our dataset is inherently very sparse. The lexicon lookup approach \cite{Sharma:2015} didn't perform well owing to the heavily misspelt words in the text, which led to incorrect transliterations as shown in Table \ref{fig:transliteration}.

\subsection{Validation of proposed hypothesis}
We obtain preliminary validation for our hypothesis that incorporating sub-word level features instead of characters would lead to better performance. Our Subword-LSTM system provides an F-score of 0.658 for our dataset, which is significantly better than Char-LSTM which provides F-score of 0.511. 

Since we do not have any other dataset in Hi-En code-mixed setting of comparable to other settings, we performed cross-validation of our hypothesis on SemEval'13 Twitter Sentiment Analysis dataset. We took the raw tweets character-by-character as an input for our model from the training set of ~7800 tweets and test on the SemEval'13 development set provided containing 1368 tweets. The results are summarized in Table \ref{tab:results_codemix}. In all the cases, the text was converted to lowercase and tokenized. No extra features or heuristics were used. 

\begin{figure*}[h]
  \captionsetup{justification=centering}
  \includegraphics[width=\textwidth]{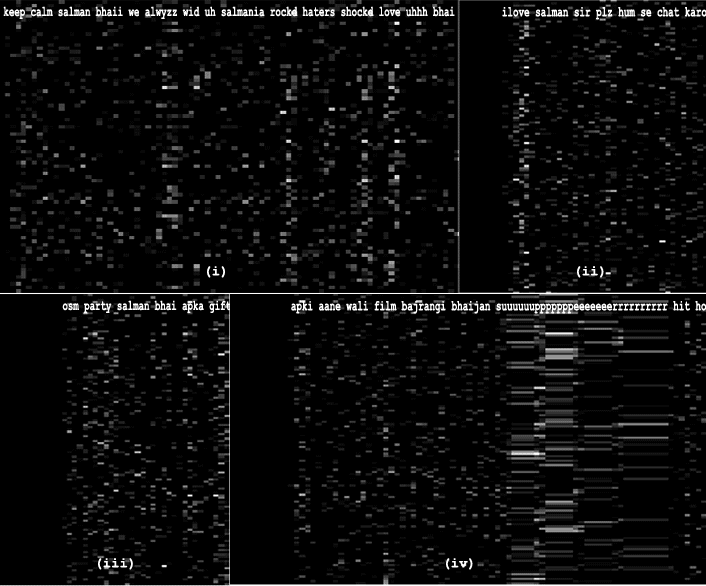}
  \caption{Visualization of the convolution layer for examples comments from the dataset show that word segments convey sentiment information despite being severely misspelt.  }
  \label{fig:visual}
\end{figure*}

\subsection{Visualizing character responses}
Visualizations in Figure \ref{fig:visual} shows how the proposed model is learning to identify sentiment lexicons. We see that different filters generally tend to learn mappings from different parts, interestingly showing shifting trends to the right which maybe due to LSTM picking their feature representation in future time steps. The words sections that convey sentiment polarity information are captured despite misspelling in example (i) and (ii). In example (iii), starting and ending phrases show high response which correspond to the sentiment conveying words (party and gift). The severe morpheme stretching in example (iv) also affects the sentiment polarity.

\section{Conclusion}

We introduce Sub-Word Long Short Term Memory model to learn sentiments in a noisy Hindi-English Code Mixed dataset. We discuss that due to the unavailability of NLP tools for Hi-En Code Mixed text and noisy nature of such data, several popular methods for Sentiment Analysis are not applicable. The solutions that involve unsupervised word representations would again fail due to sparsity in the dataset. Sub-Word LSTM interprets sentiment based on morpheme-like structures and the results thus produced are significantly better than baselines. 

Further work should explore the effect of scaling of RNN and working with larger datasets on the results. In the new system, we would like to explore more deep neural network architectures that are able to capture sentiment in Code Mixed and other varieties of noisy data from the social web.

\bibliographystyle{acl}
\bibliography{coling2016}


\end{document}